\documentclass{article}

\usepackage[utf8]{inputenc} 
\usepackage[T1]{fontenc}    
\usepackage{hyperref}      
\usepackage{url}            
\usepackage{booktabs}       
\usepackage{amsfonts}       
\usepackage{nicefrac}       
\usepackage{microtype}      
\usepackage{lipsum}		
\usepackage{graphicx}
\usepackage{natbib}
\usepackage{doi}
\usepackage{amsmath}
\usepackage{algorithm}
\usepackage{algorithmic}
\usepackage{bbm}

\usepackage{arxiv}

\title{$Q$-learning with Online Random Forests}

\date{}	

\author{
    \href{https://orcid.org/0000-0002-5541-5014}{\includegraphics[scale=0.06]{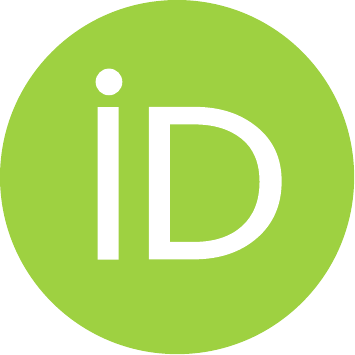}\hspace{1mm}Joosung Min} ~~~~
    \href{https://orcid.org/0000-0003-2187-7314}{\includegraphics[scale=0.06]{figs_etc/orcid.pdf}\hspace{1mm}Lloyd T. Elliott}\vspace{0.5em}\\
    Department of Statistics and Actuarial Science \\
    Simon Fraser University, 
    British Columbia, Canada \\
    \href{mailto:joosung\_min@sfu.ca}{\texttt{joosung\_min@sfu.ca}}, \href{mailto:lloyd\_elliott@sfu.ca}{\texttt{lloyd\_elliott@sfu.ca}}
}
\vspace{1mm}
\hypersetup{
pdftitle={\texorpdfstring{$Q$-learning with Online Random Forests}{}},
pdfauthor={Joosung Min, Lloyd T. Elliott},
pdfkeywords={Reinforcement learning, Online random forests},
}

\begin{document}
\maketitle

\begin{abstract}
	$Q$-learning is the most fundamental model-free reinforcement learning algorithm. Deployment of $Q$-learning requires approximation of the state-action value function (also known as the $Q$-function). In this work, we provide online random forests as $Q$-function approximators and propose a novel method wherein the random forest is grown as learning proceeds (through expanding forests). We demonstrate improved performance of our methods over state-of-the-art Deep $Q$-Networks in two OpenAI gyms (`blackjack' and `inverted pendulum') but not in the `lunar lander' gym. We suspect that the resilience to overfitting enjoyed by random forests recommends our method for common tasks that do not require a strong representation of the problem domain. We show that expanding forests (in which the number of trees increases as data comes in) improve performance, suggesting that expanding forests are viable for other applications of online random forests beyond the reinforcement learning setting. 
\end{abstract}


\section{Introduction}

In reinforcement learning (RL), agents learn to make good decisions through interaction with their environment. Such methods are used in object tracking, games, and recommendation systems and often involve online learning in which observations arrive with volume and variety. Online random forests provide lightweight implementations suitable for such data~\cite{saffari2009line}. In $Q$-learning~\cite{sutton1988learning}, the action-value function may be approximated by an arbitrary function. Variational methods~\cite{lars19}, linear function approximation methods~\cite{barto1989learning}, radial basis function approximation~\cite{powell1987radial}, and neural networks have long been a standard for functional approximation in $Q$-learning~\cite{franccois2018introduction}. In this work, we explore online random forests (ORFs; \cite{saffari2009line}) for approximation of the $Q$-function. In order to operationalize ORFs for approximation of $Q$-functions, we solve two theoretical issues: 1) We bring methods from multiple output random forests \cite{ernst2005tree} to ORFs, and 2) Previous work in ORFs is limited to categorical output, we extend this to regression trees so that the continuous $Q$-function can be approximated. We also introduce an \emph{expanding trees} method to the ORF cannon wherein the number of trees used in random forest regression begins small when the first data points come in and is increased as more data comes in (the new trees are centred at previously learned trees). 

\begin{figure*}[ht!]
    \centering
    \includegraphics[width=0.7\linewidth]{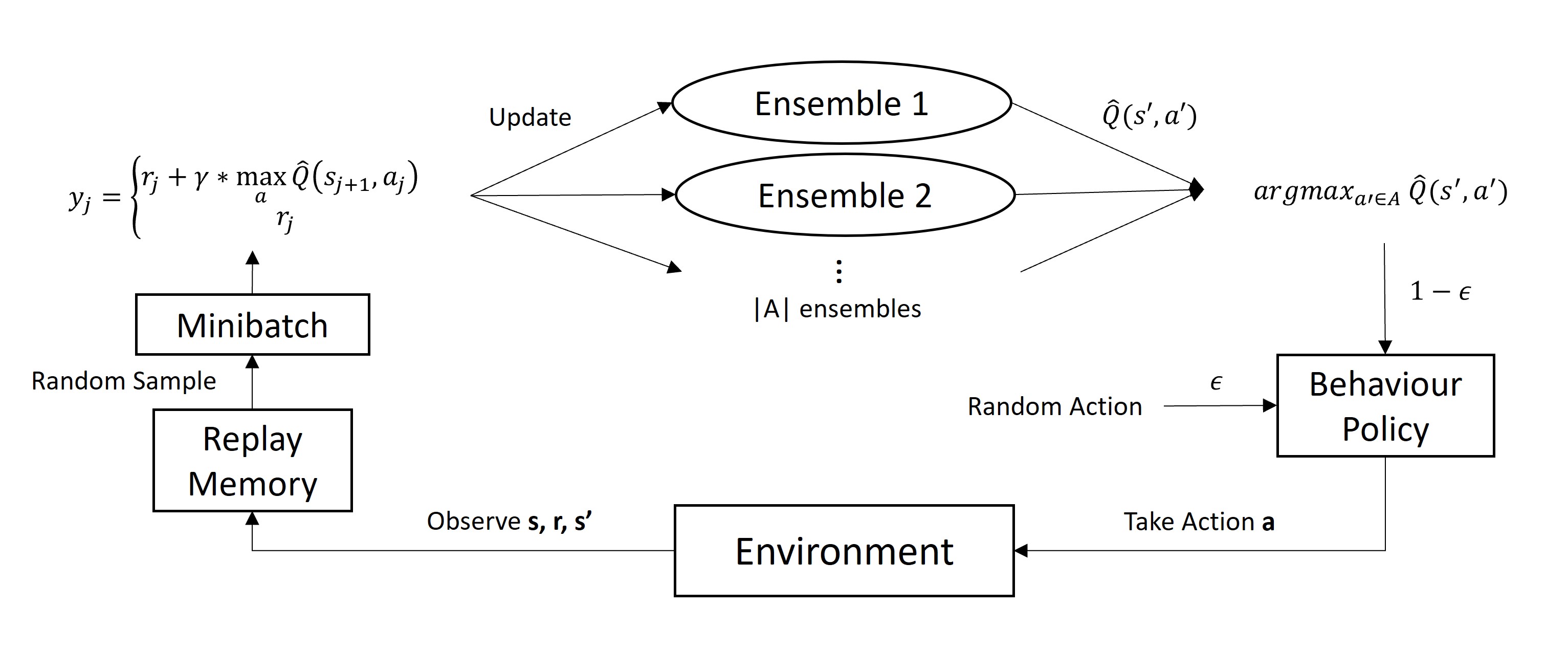}
    \caption{Schematics of $Q$-learning with online random forests. Transitions from interactions with the environments are stored in replay memory. Random minibatches of the transitions sequentially update the online trees, which approximate action-value functions for the succeeding state-action pairs. The behaviour policy chooses the action with the largest $Q$-function with the probability of 1-$\varepsilon$, or a random action with $\varepsilon$. See Section 3 for detail.}
    \label{fig:1}
\end{figure*}

In Figure~\ref{fig:1}, we provide an illustration of our methods. When an agent executes an action, the corresponding tuple of the states, action and reward is stored in replay memory. A random batch of the memory is then used for updating the online trees according to the tuples in the batch. When the agent chooses an action for a given state, the ensembles of online trees predicts the action-value functions. The agent selects the action that returns the maximum predicted reward according to the $Q$-functions with a probability of 1-$\varepsilon$, or a random action with a probability of $\varepsilon$. As the number of episodes increases, we use an `expanding trees' method to add additional trees to the ensemble. Details are provided in Algorithm S4 in the Supplementary Material.

We apply our methods to several OpenAI gym environments: blackjack, inverted pendulum, and lunar lander \cite{openaigym}, and compare to state-of-the-art Deep $Q$-Networks (DQNs) and traditional discrete temporal difference (TD) learning. We show that our version of online random forests (which we refer to as RL-ORF for \emph{reinforcement learning with online random forests}) can successfully approximate action-value functions for $Q$-learning in some gyms, with performance exceeding DQNs in the blackjack gym. In Section 2, we describe related work (including online random forests, and offline methods for tree based inference). In Section 3, we describe our methods. In Section 4 and 5, we describe our experiments on OpenAI RL gyms. In Section 6 and 7, we conclude and provide directions of future work.

\section{Related work}

We continue this section with an overview of non-linear approximation methods for $Q$-functions (including approximation through deep neural networks). Then, we discuss previous studies that implemented tree-based function approximators for RL. Lastly, we review online random forests methods and general advances in RL that we adapt for our model, including online bagging, temporal knowledge weighting, and experience replay.

\subsection{\texorpdfstring{Non-linear $Q$-function approximators and $Q$-networks}{}}
$Q$-learning is an off-policy value iteration process which requires storing and updating all possible state-action pairs and corresponding value functions~\cite{watkins1992q}.
\begin{align}
    Q(S_{t}, A_{t}) \leftarrow & Q(S_{t}, A_{t}) + \alpha [R_{t} + \gamma \max_{a \in \mathcal{A}} Q(S_{t+1}, a) \nonumber - Q(S_{t}, A_{t})].
\end{align}
Here $Q$($\cdot$) is the learned action-value function for state-action pair ($S_{t}, A_{t}$) at time $t$, $R$ is the reward, $\alpha$ is the learning rate, and $\gamma$ is the discount factor. These methods suffer from the high computational cost if there are a large number of possible states and actions and are not applicable to continuous data. Also, the action-value functions estimated in one episode cannot be generalized over the successive episodes, unless there are exact action-value matches. Approximation of the  $Q$-function allows generalization of TD learning to continuous domains, and mitigate computational costs, making the paradigm feasible. While linear approximations are fast and straightforward to implement and usually come with better convergence guarantees~\cite{barto1989learning}, they are limited as interactions between the features cannot be accounted for. These shortcomings have directed focus to non-linear function approximators, and deep neural networks ($Q$-networks) have excelled~\cite{silver2017mastering,tesauro1994td}. While $Q$-networks are effective if extensive computational power is available and deep representations of the problem domain are required, more streamlined methods for approximation of the $Q$-function (such as random forest approximations) may be indicated for problems with lower dimensional state spaces. Recent work in online random forests suggests a direction for the adoption of random forests as $Q$-function approximators. 

\subsection{Tree-based reinforcement learning}
Previous work in tree-based RL has introduced a method known as fitted $Q$-learning \cite{ernst2005tree}. In this offline method (also known as 'batch mode'), for each iteration, the algorithm builds a training set composed of observations obtained by randomly exploring the environment for a certain number of episodes as inputs. The expected reward function is induced by a supervised learning method trained using previous steps as outputs. The model is then re-trained on a training set. However, a large training set size is required to obtain good approximations, which causes high computational costs, and a tree-based ensemble must be re-built at each iteration, confining the algorithm to batch scenarios~\cite{barreto2014tree}. Silva et al.\ introduced a combination of differentiable decision trees (DDT) with policy gradient method, which outperformed an MLP in various tasks~\cite{silva2020optimization}. However, policy gradient methods often converge on a local maximum rather than the global optimum, unlike $Q$-learning which always try to reach the maximum~\cite{Sutton1998rein}. 

\subsection{Online bagging and online decision trees}
Our methods make reference to previous work in online random forests. Online random forests work by combining online bagging with extremely randomized forests \cite{saffari2009line}. The tree in the online random forest starts with a single terminal node. When new data is observed, each tree takes in the new data according to a random integer drawn from Poisson distribution. The terminal node of the tree performs splits only when the statistics computed from a series of new data exceeds a certain threshold. Trees are replaced by a new tree based on their out-of-bag errors (OOBEs). 

In \cite{saffari2009line} online bagging \cite{oza2001online} is utilized to grow trees in a non-recursive manner: as new training data is observed, include each data item a number of times for each tree, and update the trees accordingly. In the online setting, the label proportions at each terminal node are collected over time. To determine when nodes need to split further, two things need to be specified: 1) The sample size that each terminal node needs to observe to produce a robust set of statistics 2) A threshold for the information gain that produces a split that makes a good prediction. Saffari et al.~\cite{saffari2009line} proposes two new parameters: The number of samples a terminal node has to observe ($\eta$), and the threshold of gain that a split has to achieve ($\beta$). In this work, a split proceeds only if the number of observations that a paricular node has observed is greater than $\eta$. After splitting, $\mathbf{p_{jlh}}$ and $\mathbf{p_{jrh}}$ are passed on to the newly generated left and right children nodes, respectively. This method allows the new terminal nodes to participate in making predictions as soon as new online data comes in. The algorithms for creating and updating nodes are shown below and Algorithm S3 in the Supplementary Material.

\begin{algorithm}[h]
\caption{updateNode(j, $\langle x, y \rangle$)}
\begin{algorithmic}[1]
\REQUIRE Number of training samples observed: $|D_{j}|$, a set of randomly created tests in node $j$: $\textbf{H}_{j}$
\STATE $|D_{j}| += 1$
\STATE Update $\textbf{p}_{j} = \{p_{1}, \ldots, p_{k} \}$ / $|D_{j}|$ where $p_{i}$ = Number of times the label $i$ appeared in $j$
\STATE Update $\textbf{p}_{jlh}$ and $\textbf{p}_{jrh} \forall h \in \textbf{H}_{j}$
\STATE Compute $\Delta L(D_{j})$
\IF{$|D_{j}| > \eta$ and $\exists h_{j} \in H_{j}: \Delta L(D_{j}, h_{j}) > \beta$}
    \STATE $h_{split} = \text{argmax}_{h \in H} \Delta L(D_{j}, h)$
    \STATE $\text{createChild}(\textbf{p}_{jlh_{split}})$ \# create left child node
    \STATE $\text{createChild}(\textbf{p}_{jrh_{split}})$ \# create right child node
\ENDIF
\end{algorithmic}
\label{alg:orf_updateNode}
\end{algorithm}
\vspace{-0.5em}

\subsection{Experience replay and temporal knowledge weighting}

Let $c$ be the number of times an observation is used in a bootstrapped sample in an online random forest. Temporal knowledge weighting developed by Saffari et al.~\cite{saffari2009line} uses observations with $c = 0$ to estimate the trees' out-of-bag error (OOBEs), and discards trees with large OOBEs. To prevent trees from being discarded in their early stages of growing, Saffari et al.~\cite{saffari2009line} also employs another parameter $\varphi$; the temporal knowledge weighting rate. Only trees with $\text{age}_{t} > 1/{\varphi}$ are subjected to being discarded. Here, $\text{age}_{t}$ is the number of samples a tree has observed. The tree to be discarded is then randomly chosen and replaced by a new tree with just one node (a stump). While the influence of discarding one tree in an ensemble is relatively low, continually replacing the trees allows the ensemble to adapt to the changes in the sample distribution throughout time~\cite{saffari2009line}. 

Another breakthrough in deep $Q$-learning is the implementation of \textit{experience replay}~\cite{lin1992self}, and this is where the acclaimed method used in \cite{mnih2013playing} differs from the original TD-gammon~\cite{tesauro1994td}. At each step $t$, the agent interacts with the environment, and outputs a tuple of experience $e_{t} = (s_{t}, a_{t}, r_{t}, s_{t+1})$ (here $s$ is the state, $a$ is the action and $r$ is the reward), and this tuple is stored in a dataset $\mathbbm{D} = e_{1}, e_{2}, \dots , e_{N}$ referred to as \textit{replay memory} \cite{mnih2013playing}. Only the most recent $N$ observations are stored. When updating the $Q$-values, we use a fixed-sized minibatch drawn at uniform random from $\mathbbm{D}$. Then, the agent chooses the next action to execute by an $\varepsilon$-greedy policy from $Q(s_{t+1}, A)$. Utilization of experience replay has several advantages over learning directly from the most recent experience only. The method is more data-efficient since each observation participates in updating the $Q$-function multiple times instead of being thrown away after being used only once. Also, using randomly drawn samples mitigates correlation between the consecutive observations, reducing the prediction variance of the $Q$-network during $Q$-value updates. Finally, experience replay prevents the behaviour policy from getting stuck in local modes. For instance, without experience replay, if an action to move left is executed, the next consecutive inputs for $Q$-value updates would occur from the left side. Then, when the action to move right is chosen, the samples would shift to the right. This phenomenon can cause the parameters of the $Q$-network to get stuck or even diverge \cite{mnih2013playing}.
Deep neural network (DNN) methods involving the above breakthroughs in RL have drawn public attention after Alpha Go defeated a human Go champion Sedol Lee in 2016 \cite{lee2016human}. RL is now widely used in solving real-world problems such as object detection, and recommendation systems  \cite{franccois2018introduction,lee2016human,Sutton1998rein,sze2017efficient}.

\section{Methods} 

Our method is as follows: at each time step, an action is taken by the agent, and the corresponding four-tuple $(s_t,a_t,r_t,s_{t+1})$ is observed from the environment, which is saved in the replay memory. Each tree in the ensemble updates either its terminal node or out-of-bag-error (OOBE) depending on an integer drawn from the Poisson distribution. The OOBE we use is described in the next subsection. Trees are grown until their age reaches 1/$\varphi$, where they become subject to being replaced according to their OOBEs. At episode $\delta$, the tree with the lowest OOBE is duplicated several times so that the ensemble size is expanded to $|M_{max}|$. The episode terminates when the agent reaches the terminal state of the environment. An overview of our method is given in Algorithm S1 in the Supplementary Material. 

\subsection{Online random forests in regression settings}

The seminal work in original online random forests~\cite{saffari2009line}, focuses on classification problems. Generalization of this work to regression is required for RL, as the $Q$-function is continuous. We achieve this generalization by replacing the objective of splits from maximizing the information gains to maximizing the change in residual sum of squares (RSS), as is done in the batch mode regression tree methods.

In addition, the computation of out-of-bag errors (OOBEs) of the trees must be computed in a different way for regression. In classification, the OOBE for a tree is simply the fraction of the new observation's label $y_{u}$ for some $u$ $\in k$ in node $j$: $\text{OOBE}_\text{class.} = (\sum_{i=1}^{|D_{j}|}  \mathbbm{1} (y_{i} \neq y_{u}))/{|D_{j}|}$. This quantity naturally falls in the interval [0, 1]. However, for regression, this quantity is not necessarily between zero and one, and so we develop a normalized mean absolute error. In our method, the OOBE is computed based on the $\lambda$ most recent observations that a node has seen, according to the following equation:

\vspace{-0.5em}
\begin{equation}
    \text{OOBE}_\text{reg.} = \frac{1}{\lambda} \sum_{i=1}^{ \lambda } \text{min}\left(\text{abs}\left(\frac{y_{i} - m(x_{i})}{y_{i} + \mu}\right), 1\right).
    \label{eq:oobe}
\end{equation}
Here $m(x)$ is the predicted response from the tree, and $\mu$ is a small arbitrary real number added for numerical stability. The operand of the $\text{min}(\cdot)$ function tells us how much the predicted value is off from the true response, as a value between [0 ,1]. 

\subsection{\texorpdfstring{Computing $\textmd{max}_{a\in \mathcal{A}} \hat Q(S,a)$ when $|A| > 1$}{}}

In reinforcement learning, there is usually more than one possible action afforded by any given non-terminal state. For agents to determine which action to choose, each action needs its own action-value. This gives rise to the need for the function approximator to be able to produce a number of outputs equal to the number of available actions. Deep neural networks naturally solve this problem by having the corresponding number of output nodes in the output layer. However, the online random forest from~\cite{saffari2009line} can approximate only one output. To resolve this, we adopt an idea presented in \cite{ernst2005tree} for handling discrete action spaces. In our method, we grow one forest for each action available in a given reinforcement learning environment. Each ensemble starts with just one node and grows independently on sample observations from experience replay. Each forest approximates the corresponding action-value, and then the largest among them is returned. For a fixed state $s$ and action $a$ at time step $t$,
\begin{align}
    \max_{a \in \mathcal{A}} \hat Q(s_{t+1}, a) = \max[ & \hat Q_{1}(s_{t+1}, a_{1}), \hat Q_{2}(s_{t+1}, a_{2}), \dots , \hat Q_{|A|}(s_{t+1}, a_{|A|}) ].
\label{eq:approx}
\end{align}
Here, $\hat Q(\cdot)$ is the function approximator, and $\hat Q_{i}(s_{t+1}, a_{i}) = M_{i}(s_{t+1}) \space \forall i = 1,\ldots,|A|$ where $M(\cdot)$ denotes prediction by the ensemble. An equivalent method is also applied for action selections: with probability $\varepsilon$, the agent chooses its consecutive action by taking the largest action-value among the approximations induced from the ensembles. This is encoded in Algorithm S2 in the Supplementary Material.

\begin{figure*}[ht!]
    \centering
    \includegraphics[width=0.49 \linewidth]{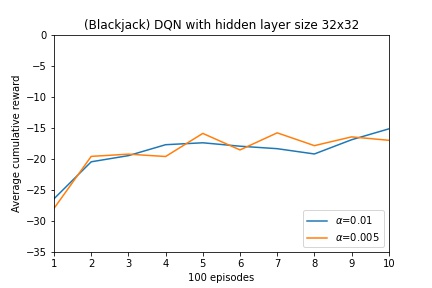}
    \includegraphics[width=0.49 \linewidth]{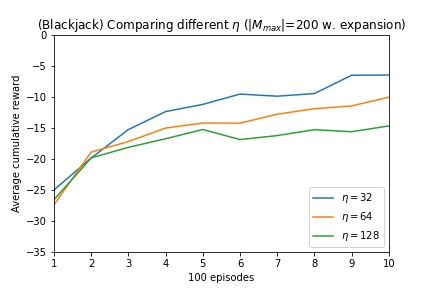}
    \caption{Results on blackjack using DQN and RL-ORF as $Q$-function approximators. (Upper) DQN with hidden layer size 32x32 with $\alpha = 0.05$ showed the highest average cumulative reward at episode 1,000. (Lower) RL-ORF with $\eta$ = 32 and ensemble size expansion returned the best average cumulative reward, exceeding DQN's performances.}
    \label{fig:2}
    \vspace{-1em}
\end{figure*}

\subsection{Expanding ensemble size}

Growing a large number of trees in each forest can be computationally expensive. In the early stages of learning, the amount of information learned may be expressed using only a small number of trees. We propose training a small number of trees for a certain period and then duplicating the best tree a large number of times later on  in the learning process. Our method begins the learning process with just 100 trees in each forest, and expands the number of trees up to a value specified by $|M_{max}|$ according to Algorithm S4 in the Supplementary Material.

\subsection{Partial randomness in node splits}
In the original online random forest from~\cite{saffari2009line}, each node selects a subset of features at random. However, it is essential in reinforcement learning that the function approximators utilize as much information from the state as possible. For this reason, we remove randomness in the split variable selection but leave split points selection for each test to be still drawn at random (hence \textit{partial randomness}). This process is described in Algorithm S3 in the Supplementary Material.

\section{Experiments}

We apply the RL-ORF model to the OpenAI blackjack, inverted pendulum, and lundar lander gyms. In each case, we compare our methods to DQNs. In OpenAI's blackjack environment, the state is a tuple containing three elements: the agent's hand, the dealer's hand, whether or not the agent holds an ace. The ace can be treated as either one point or eleven. The agent starts with two cards in hand, whereas the dealer starts with only one. The player draws a card by choosing to `hit' (for more details, see \cite{openaigym}). Each epoch is one thousand episodes, and each episode consists of multiple time steps (extending over the length of the simulated task). For each time step in the episode, the agent executes an action selected by the behaviour policy and the resulting transition is stored in the replay memory. Random samples are drawn from the memory and the samples are used to update the trees according to Algorithm~\ref{alg:orf_updateNode} and Algorithm S3 in the Supplementary Material. For both DQN and RL-ORF, we used $\gamma$ = 1, $\varepsilon$ = 0.5, and $\varepsilon$-decay = 0.99 with the minimum $\varepsilon$ = 0.01. Throughout our experiments, the replay memory size was 10,000 in all experiments with a minibatch size of 32. For DQN, the neural networks comprised two hidden layers. In each experiment, we compared performances of different hidden layer sizes and learning rates with Adam optimizer \cite{kingma2014adam}. The input and output layer sizes differed depending on the environment. For RL-ORF, we tested different values of $\eta$ and whether to expand the ensemble size or not. All trees were fully grown without pruning, along with $\beta = 0.01$, $\varphi = 1/5,000$, $|M_{init}|$ = 100, and $|M_{max}|$ = 200. The average rewards per epoch are summed and shown in units of 100 episodes. For each experiment, we do 100 random restarts, 1,000 episodes for each run. 

For the `inverted pendulum' by OpenAI's CartPole-v1, the agent's objective is to maintain the pole standing on the cart without falling as long as possible. The state-space tuple consists of 4 elements: cart position, cart velocity, pole angle, and pole angular velocity. There are two possible actions: move the cart to the left (0) or right(1). The agent gets +1 reward for every step taken, including the termination step \cite{openaigym}. To enhance learning speed, we modified the reward function for both DQN and RL-ORF. In the altered setting, the agent gets a -1,000 reward for falling. We tested hidden layer sizes \{32x32, 64x64, 128x128\}, and learning rates($\alpha$) \{0.01, 0.005\} for DQN. For RL-ORF, different values of $\eta$=\{32, 64, 128\}, and whether to expand the ensemble size from 100 to 200.

Finally, for the `lunar lander' gym the agent tries to land on the landing pad located at coordinates (0,0). The state-space tuple consists of 8 elements: x-coordinate, y-coordinate, horizontal and vertical velocity, lander angle, angular velocity, right-leg grounded, and left-leg grounded. The agent gets -0.3 points for each frame it fires the main engine, -0.03 for each side engine. If the lander reaches the ground too fast (speed $>$ 0), the lander crashes and receives -100. A successful landing (velocity = 0) anywhere awards +100 points, an additional +100 are given for landing on the landing pad. Each leg with ground contact is +10 points. An episode terminates when the lander either crashes or comes to rest \cite{openaigym}. We demonstrate our results for DQN with hidden layer size 32x32, $\alpha$= 0.01, and for RL-ORF with $\eta = 256$ and expand ensemble sizes to $|M_{max}|$ = 200.

All experiments were conducted on Intel i7-8565U 1.8GHz CPU and 16GB RAM with python version 3.7.8. The DQN is trained using \emph{PyTorch} version 1.8.1. The code for our experiments are available under an open source license. Portions of our codebase use a modified version of the open source python code from \cite{luiarther} and \cite{liuyuxi}. 

\section{Results}
\subsection{Experiment 1: Blackjack}

\begin{figure*}[ht!]
    \centering
    \includegraphics[width=0.49 \linewidth]{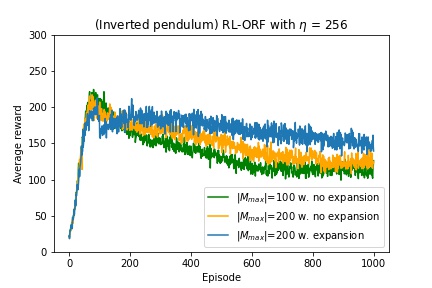}
    \includegraphics[width=0.49 \linewidth]{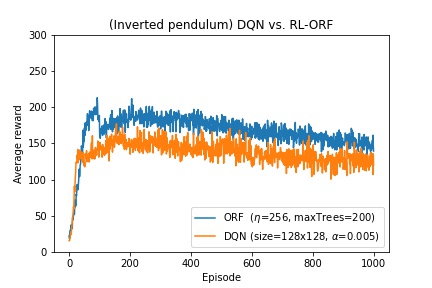}
    \caption{Results on the inverted pendulum environment. (Left) expanding the number of trees showed better performance and a slower rate of catastrophic forgetting. (Right) RL-ORF with $\eta = 256$ and ensemble size expansion outperformed DQN size 128x128 with $\alpha = 0.005$ at episode 1,000.}
    \label{fig:3}
\end{figure*}

In Figure~\ref{fig:2}, we show representative performance of the DQN and the RL-ORF over a number of hidden layer sizes and parameter settings (figures depicting the full range of parameters explored for each method are provided in Figures S1, S2 and S3  of the Supplementary Material). The settings for the RL-ORF that performed best are $\eta$ = 32 and ensemble size expansion from 100 to 200 at episode 100, and the settings of the best performing DQN are hidden layer size 32x32 with $\alpha = 0.05$. We varied the parameters of the DQN extensively but found that the settings did not modulate the performance (Figure S1 of the Supplementary Material). For this gym, the RL-ORF performed significantly better than the DQN. The best RL-ORF parameters showed a mean cumulative reward of -6.51 with a standard deviation of 10.97 at episode 1,000, and the best DQN had a mean of -15.18 with a standard deviation of 10.10. A one-sided t-test for difference in the means yields the $p$-value 2.271E-5 (more detail for this test is provided in Table S1 of the Supplementary Material), rejecting the null hypothesis that the performances were equal (i.e., the RL-ORF has higher mean performance). A Shapiro-Wilk test provides no evidence that the distributions of the performances are not normal (this test is described in Table S1 of the Supplementary Material).

Note that the formulation of blackjack by OpenAI reshuffles the deck after every hand, and ties are not in favour of the agent. This means that the game is stacked against the agent, and it is impossible to achieve average reward higher than zero (as indicated by Figure~\ref{fig:2}). For the blackjack experiment, there was no evidence of catastrophic forgetting in the DQN (in Figure~\ref{fig:2}, Left, the DQN performance does not decrease after plateau). Also, in this experiment there was no evidence that the \emph{expanding trees} method improved performance of the RL-ORF (Figure 2 of the Supplementary Material). Evidence recommending \emph{expanding trees} arises in the next experiment. We also apply standard $Q$-learning (discrete TD learning) to the blackjack gym, and this method performed worse than both the DQN and the RL-ORF methods, with a mean reward of -28.93 and a standard deviation of 12.97 (averaged over 100 random restarts).

\subsection{Experiment 2: Inverted pendulum}
For the inverted pendulum, the RL-ORF performed significantly better than the DQN in some cases, as shown in Figure~\ref{fig:3} (this Figure shows the learning using the best parameter settings found for both methods). With the RL-ORF settings $\eta = 256$ with ensemble size expansion from 100 to 200 at episode 100, and DQN settings $\alpha = 0.005$ with hidden layer size 128x128, a Mann-Whitney U-test \cite{mann1947test} rejects the null hypothesis (with a $p$-value of 0.009) that the average reward is the same for RL-ORF and DQN at episode 1,000 (a Shapiro-Wilk test for normality shows that the rewards for this experiment are not normal and so we prefer the Mann-Whitney U-test over the t-test: details for these tests are provided in Tables S3 and S4 of the Supplementary Material). The average reward at the 1,000th episode for DQN is 120.04$\pm$91.99 and the average reward for RL-ORF is 139.26$\pm$88.78, indicating that RL-ORF is better. In addition, for this gym we find that our expanding trees method improves the RL-ORF performance. Comparisons of all of the parameter settings (beyond  Figure~\ref{fig:3}) are provided in Figures S4, S5 and S6 of the Supplementary Material, including error bars.

We found that in this gym, a lower learning rate of $\alpha = 0.005$ gave better performance than $\alpha = 0.01$ for the DQN. However, like in blackjack, a larger number of hidden nodes did not meaningfully improve the overall performance (see Figure S4 of the Supplementary Material). Finally, in both DQN and RL-ORF, we  see that the average reward slowly decreases over the episodes. The problem is often referred to \emph{catastrophic forgetting} \cite{mccloskey1989catastrophic}. After reinforcement learning algorithms achieve a reasonable solution to a problem, new incoming experiences that the agent gets are only for `good' cases, leading to a depletion of unsuccessful cases in the experience memory. The function approximator can then start to generate high $Q$-function values for every state-action pair, which degenerates the accuracy of the agent.

In the lunar lander experiment, neither the DQN nor the RL-ORF performed well: neither method could successfully land the lunar lander a single time over thousands of epochs (this is displayed in Figure S7 in the Supplementary Material). While not managing to land, the DQN method provided more efficient fuel use before cratering, leading to improved average reward over RL-ORFs.

\section{Discussion and future work}

We discovered that the online tree method could outperform some deep neural networks in terms of average total reward. In the process, we found that starting the forest size with a small number of trees and then expanding the size after the set episodes performed better in later episodes. This may be because when the forest size is expanded, more than half of the trees temporarily show relatively high performance, which could cancel out the performance degradation due to correlations between the trees and tree re-growth. The impact of tree expansion decreases as $\eta$ increases, indicating that the ensemble becomes more robust to changes with larger $\eta$. However, online tree methods did not perform well in more complicated environments such as the lunar lander. We believe it would be worth investigating what makes it difficult for the online tree methods to solve those problems. Another limitation is that our online tree method is coded entirely in Python, which made the learning process around 100 times slower than the DQN built on PyTorch. We believe parallelization and implementation of our method in lower-level languages such as C would boost the learning speed.

\section{Conclusion}

We have developed an online random forest method for reinforcement learning. This method is general and we apply it to gyms without any hand-crafted aspects, without transfer learning, and without building specific representations of the gym. Our experiments demonstrate that we outperform state-of-the-art DQNs and standard TD learning for blackjack and we outperform DQNs in the inverted pendulum.

These gains come at a cost: our method is significantly slower than DQNs (however, our DQN implementation uses \emph{torch} and we did not attempt to optimize our code with a C implementation matching \emph{torch} optimization techniques). The lunar lander gym is quite difficult and neither DQN nor our method performs well for that gym. The lunar lander gym would likely be easier with visual representation and convolution, or hand-crafted representations. In addition to providing the RL-ORF (reinforcement learning online random forest), our work shows some limitations on the complexity of problems that can be solved with representation-free reinforcement learning.

\end{document}